\documentclass{article}

\usepackage{microtype}
\usepackage{graphicx}
\usepackage{subcaption}
\usepackage{booktabs} % for professional tables

\usepackage{hyperref}

\usepackage[accepted]{icml2026}

\usepackage{amsmath}
\usepackage{amssymb}
\usepackage{mathtools}
\usepackage{amsthm}

\usepackage[capitalize,noabbrev]{cleveref}

\theoremstyle{plain}

\theoremstyle{definition}

\theoremstyle{remark}

\usepackage[textsize=tiny]{todonotes}

\icmltitlerunning{JI-ADF: Joint-Individual with Adaptive Decision Fusion}

\begin{document}

\twocolumn[
  \icmltitle{JI-ADF: Joint-Individual Learning with Adaptive Decision Fusion for Multimodal Skin Lesion Classification}



  \icmlsetsymbol{equal}{*}

  \begin{icmlauthorlist}
    \icmlauthor{Phan Nguyen}{equal,kaist}
    \icmlauthor{Dat Cao}{equal,kaist}
    \icmlauthor{Hien Kha}{tmu}
    \icmlauthor{Hien Chu}{kaist}
    \icmlauthor{Minh Le}{yale} 
    \icmlauthor{Trang Pham}{tmu}
    \icmlauthor{Nguyen Quoc Khanh Le}{tmu}
  \end{icmlauthorlist}

  \icmlaffiliation{kaist}{KAIST}
  \icmlaffiliation{tmu}{Taipei Medical University}
  \icmlaffiliation{yale}{Yale University}

  \icmlcorrespondingauthor{Phan Nguyen}{nhphan@kaist.ac.kr}
  % \icmlcorrespondingauthor{Nguyen Quoc Khanh Le}{khanhlee@tmu.edu.tw}

  \icmlkeywords{Machine Learning, ICML}

  \vskip 0.3in
]



\printAffiliationsAndNotice{}  % no special notice (required even if empty)

\begin{abstract}
Skin lesion classification is essential for early dermatological diagnosis, yet many existing computer-aided systems rely primarily on dermoscopic images and underutilize the multimodal evidence routinely available in clinical practice. To address this gap, we propose \textbf{JI-ADF}, a trimodal deep learning framework that integrates dermoscopic images, clinical photographs, and structured patient metadata for clinically grounded skin lesion classification. The proposed architecture combines joint multimodal representation learning with modality-specific auxiliary supervision and an adaptive decision fusion mechanism that dynamically calibrates modality contributions on a per-sample basis. To enhance cross-modal reasoning while preserving modality-specific evidence, we further introduce a multimodal fusion attention (MMFA) module. We evaluate JI-ADF on the large-scale MILK10k benchmark. Extensive analyses, including modality ablation, calibration evaluation, and Grad-CAM visualization, further confirm the robustness and clinically meaningful behavior of the model. The results indicate that JI-ADF provides a reliable and practical foundation for multimodal skin lesion classification in real-world clinical settings.
\end{abstract}

\section{Introduction}
\label{sec:intro}

Skin cancer is one of the most prevalent malignancies worldwide, with melanoma accounting for a disproportionate number of deaths. While the 5-year survival rate for localized melanoma is $99\%$, it drops to approximately $35\%$ at metastatic stages, making early diagnosis critical \cite{bray2024globalcancer2022, acs_cancerfacts2024, ferlay2024cancertomorrow}. Deep learning has significantly advanced skin lesion classification, with CNN-based models reaching dermatologist-level accuracy on ISIC benchmarks in some studies \cite{7792699, 9018274, 8990108, Esteva2017, Haenssle2018}. However, most existing systems rely solely on dermoscopic images, limiting the ability to reflect real clinical practice where diagnosis depends on both visual and contextual information.

Recent work has explored multimodal learning to integrate images with clinical context and metadata \cite{Atrey2010, Huang2020}. In dermatology, combining dermoscopic images, clinical photographs, and patient information has been shown to improve classification performance \cite{Yap2018, Liu2020, 9364366, BI2020107502}. Nevertheless, many approaches rely on simple concatenation or late fusion, which fail to capture complex cross-modal interactions \cite{PACHECO2020103545, 9098645}. While attention-based methods partially address this issue \cite{9364366, Cai2023}, most designs still compress multimodal information into a single representation and treat metadata primarily as auxiliary guidance, limiting their ability to model complementary and instance-specific modality contributions \cite{9363019, 8027090}.

To address these limitations, we propose JI-ADF, a trimodal framework that integrates dermoscopic images, clinical images, and structured metadata through attention-based interaction and adaptive decision fusion. Our contributions are as follows:
\begin{itemize}
    \item We propose JI-ADF, a trimodal framework that jointly models dermoscopic images, clinical images, and structured metadata through attention-based interaction.
    \item We introduce an adaptive decision fusion strategy that dynamically combines joint and modality-specific predictions, improving robustness when modalities provide unequal diagnostic evidence.
    \item We conduct extensive experiments on the MILK10k benchmark, showing that JI-ADF outperforms existing multimodal approaches and achieves balanced performance across lesion types.
\end{itemize}

\begin{figure*}[ht]
    \centering
    \includegraphics[width=0.76\textwidth]{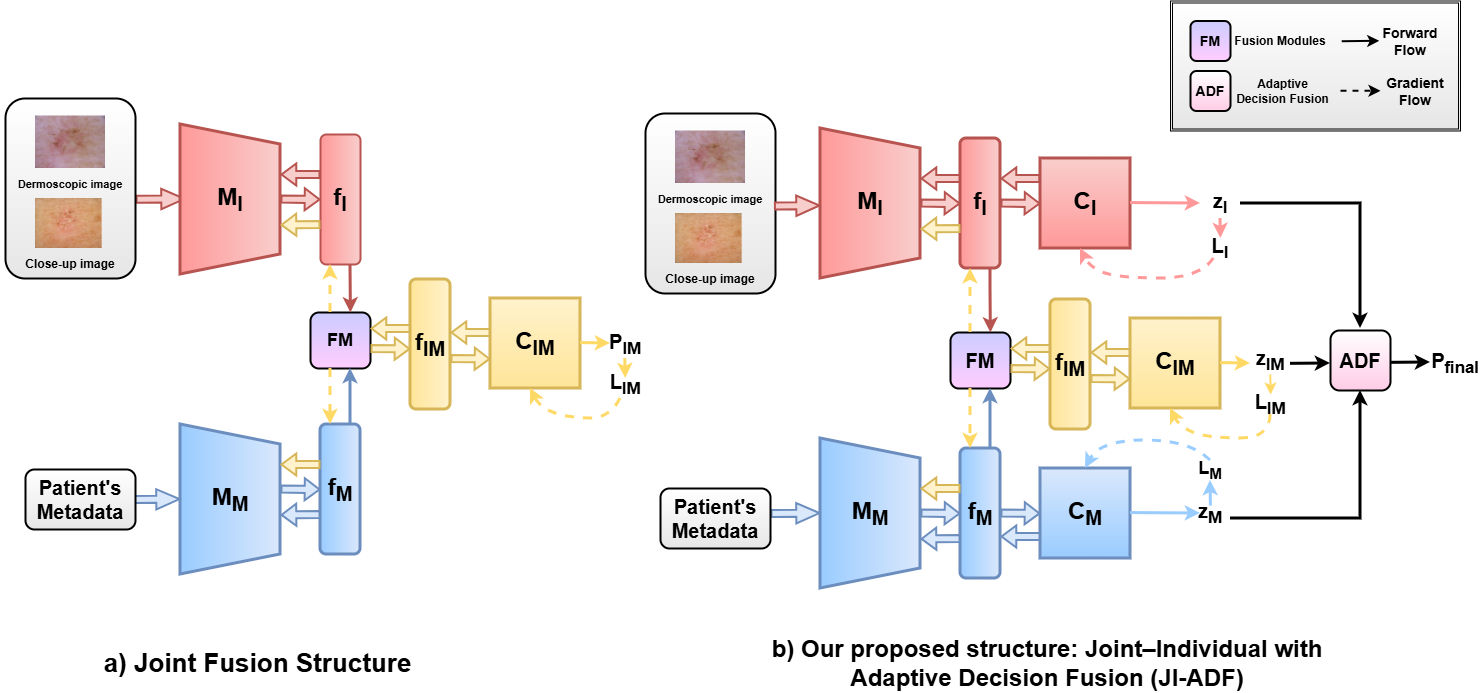}
    \caption{Illustration of (a) the Joint Fusion Structure and (b) our proposed Joint–Individual architecture with Adaptive Decision Fusion. JI-ADF extends the baseline by adding individual prediction heads for each modality and an adaptive fusion module that assigns instance-dependent weights.}
    \label{fig:2_methods}
\end{figure*}

% \input{sec/2_related}
% \begin{figure*}[!t]
%     \centering
%     \includegraphics[width=0.85\textwidth]{img/JF_ADF.png}
%     \caption{Illustration of (a) the Joint Fusion Structure and (b) our proposed Joint–Individual architecture with Adaptive Decision Fusion. JI-ADF extends the baseline by adding individual prediction heads for each modality and an adaptive fusion module that assigns instance-dependent weights.}
%     \label{fig:2_methods}
% \end{figure*}

\section{Method}

\subsection{Proposed Architecture: Joint--Individual with Adaptive Decision Fusion}

% \begin{figure}[htbp]
%  % Caption and label go in the first argument and the figure contents
%  % go in the second argument
% \floatconts
%   {fig:proposed}
%   {\caption{Our proposed method: Joint–Individual with Adaptive Decision Fusion (JI-ADF)}}
%   {\includegraphics[width=1.0\linewidth]{img/ADF.png}}
% \end{figure}

% \begin{figure*}[!ht]
%     \centering
%     \includegraphics[width=0.85\textwidth]{img/JF_ADF.png}
%     \caption{Illustration of (a) the Joint Fusion baseline and (b) our proposed Joint–Individual architecture with Adaptive Decision Fusion. JI-ADF extends the baseline by adding individual prediction heads for each modality and an adaptive fusion module that assigns instance-dependent weights.}
%     \label{fig:2_methods}
% \end{figure*}

\paragraph{Backbone streams.}
From the two images and the metadata, the modality encoders produce features
\begin{equation}
\begin{gathered}
\mathbf{f}_I = M_I\!\big(I_{\mathrm{derm}}, I_{\mathrm{close}}\big) \in \mathbb{R}^{D_I} \\
\mathbf{f}_M = M_M(M) \in \mathbb{R}^{D_M}
\end{gathered}
\end{equation}
A differentiable fusion module aggregates them into a joint representation
\begin{equation}
\mathbf{f}_{IM} = FM(\mathbf{f}_I, \mathbf{f}_M) \in \mathbb{R}^{D_{IM}}
\end{equation}

\paragraph{Branch classifiers and auxiliary supervision.}
Each stream has its own classifier head
\begin{equation}
\begin{gathered}
\mathbf{z}_I = C_I(\mathbf{f}_I) \in \mathbb{R}^{N},\ 
\mathbf{P}_I = \mathrm{softmax}(\mathbf{z}_I) \\
\mathbf{z}_M = C_M(\mathbf{f}_M) \in \mathbb{R}^{N},\ 
\mathbf{P}_M = \mathrm{softmax}(\mathbf{z}_M) \\
\mathbf{z}_{IM} = C_{IM}(\mathbf{f}_{IM}) \in \mathbb{R}^{N},\ 
\mathbf{P}_{IM} = \mathrm{softmax}(\mathbf{z}_{IM})
\end{gathered}
\end{equation}

With one-hot target $\mathbf{y}$ we use cross-entropy on all three branches
\begin{equation}
\begin{gathered}
\mathcal{L}_I = -\sum_{c=1}^{N} y_c \log P^{(c)}_I \\
\mathcal{L}_M = -\sum_{c=1}^{N} y_c \log P^{(c)}_M \\
\mathcal{L}_{IM} = -\sum_{c=1}^{N} y_c \log P^{(c)}_{IM}
\end{gathered}
\end{equation}

\paragraph{Adaptive Decision Fusion (ADF).}
Instead of a fixed average at the decision level, we learn per-sample fusion weights from the joint evidence of all heads.
Let $\mathbf{s}=[\,\mathbf{z}_I \,\|\, \mathbf{z}_{IM} \,\|\, \mathbf{z}_M\,]\in\mathbb{R}^{3N}$ be the concatenated logits, where $[\cdot \| \cdot]$ denotes vector concatenation.
A lightweight gating network produces simplex weights
\begin{equation}
\begin{aligned}
\boldsymbol{\alpha}
&= \mathrm{softmax}\!\big(W_2\,\sigma(W_1 \mathbf{s} + \mathbf{b}_1) + \mathbf{b}_2\big) \\
&= (\alpha_I,\alpha_{IM},\alpha_M)
\end{aligned}
\end{equation}
where $\sigma(\cdot)$ denotes a pointwise (element-wise) nonlinearity and $\sum_{k\in\{I,IM,M\}}\alpha_k=1$.
The final posterior is a convex combination of branch posteriors
\begin{equation}
\mathbf{P}_{\mathrm{final}}
= \alpha_I \mathbf{P}_I
+ \alpha_{IM} \mathbf{P}_{IM}
+ \alpha_M \mathbf{P}_M
\in \Delta^{N-1}
\end{equation}
We use a softmax head and take the prediction by
$\hat y=\arg\max_{c\in\{1,\dots,N\}} P_{\mathrm{final}}^{(c)}$.
\paragraph{Training objective.}
We supervise the fused prediction and the auxiliary heads
\begin{equation}
\mathcal{L}_{\mathrm{total}} =
\mathrm{CE}(\mathbf{P}_{\mathrm{final}}, \mathbf{y})
+ \lambda_{IM}\mathcal{L}_{IM}
+ \lambda_{I}\mathcal{L}_{I}
+ \lambda_{M}\mathcal{L}_{M}
\end{equation}

We fix the auxiliary weights to $\lambda_{IM}=0.5$ and $\lambda_{I}=\lambda_{M}=0.25$ throughout. This keeps the total auxiliary weight at most equal to the unit weight on the final loss, emphasizes the joint branch that is closest to deployment, and treats the individual branches as regularizers that stabilize training and preserve modality-specific cues.

\subsection{Multi-Modal Fusion Attention (MMFA)}\label{sec:MMFA}

We instantiate the fusion module \(FM\) as a \emph{multimodal fusion attention} block that lets image and metadata features attend to each other while preserving self-evidence, following prior work~\cite{TANG2024110604}.

\begin{figure}[ht]
    \centering
    \includegraphics[width=0.85\linewidth]{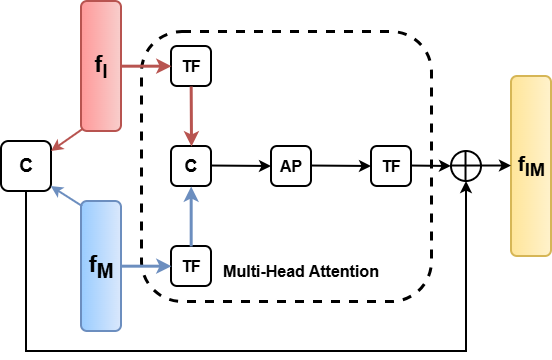}
    \caption{Multimodal Fusion Attention Module (MMFA), where image features $f_{I}$ and metadata features $f_{M}$ are jointly refined through cross-attention and self-attention mechanisms between the two modalities. The resulting fused representation $f_{IM}$ captures the enhanced joint features after integration. (\textbf{C}: Concatenation, \textbf{AP}: Attention Operation, \textbf{TF}: Transform).}
    \label{fig:mmfa}
\end{figure}

\begin{table*}[ht]
\centering
\small
\setlength{\tabcolsep}{7pt}
\begin{tabular}{@{}p{0.33\linewidth}cccccc@{}}
\toprule
\textbf{Method} &
\textbf{AUC} &
\textbf{Precision} &
\textbf{Accuracy} &
\textbf{Sensitivity} &
\textbf{Specificity} &
\textbf{Dice} \\
\midrule
JIF-MMFA \cite{TANG2024110604} & 0.750 & 0.328 & 0.896 & 0.406 & 0.939 & 0.306 \\
VEMFL \cite{restrepo2024multimodaldeeplearninglowresource} & 0.842 & 0.439 & 0.912 & 0.291 & 0.958 & 0.302 \\
CAFFM \cite{TRANVAN2025102588} & 0.787 & 0.306 & 0.898 & 0.233 & 0.954 & 0.232 \\
ALBEF \cite{Adebiyi2024.05.30.24308213} & 0.760 & 0.294 & 0.889 & 0.208 & 0.945 & 0.222 \\
CoscatNet-UFS \cite{ZUO2025103091} & 0.800 & 0.472 & 0.925 & 0.453 & \textbf{0.959} & 0.435 \\
DualRefNet \cite{Khurshid2025} & 0.848 & 0.447 & 0.918 & 0.422 & 0.954 & 0.441 \\
SkinM2Former \cite{10943957} & 0.841 & 0.436 & 0.920 & 0.421 & 0.954 & 0.405 \\
Fine-tuned PanDerm \cite{Yan2025} & 0.840 & 0.476 & 0.920 & 0.447 & 0.953 & 0.434 \\
\midrule
\textbf{JI-ADF (ours)} & \textbf{0.866} & \textbf{0.543} & \textbf{0.930} & \textbf{0.536} & \textbf{0.959} & \textbf{0.505} \\
\bottomrule
\end{tabular}
\caption{Comparison with state-of-the-art skin lesion classification approaches.}
\label{tab:comparision}
\end{table*}

\paragraph{Inputs and projections.}
Given $\mathbf{f}_I\!\in\!\mathbb{R}^{D_I}$ and $\mathbf{f}_M\!\in\!\mathbb{R}^{D_M}$, each head $h=1,\dots,H$ forms modality–specific queries, keys, and values
\begin{equation}
\begin{aligned}
\mathbf{Q}_h &=
\begin{bmatrix}
W_h^{Q,I}\mathbf{f}_I \\
W_h^{Q,M}\mathbf{f}_M
\end{bmatrix}
\in\mathbb{R}^{2\times d_h} \\
\mathbf{K}_h &=
\begin{bmatrix}
W_h^{K,I}\mathbf{f}_I \\
W_h^{K,M}\mathbf{f}_M
\end{bmatrix}
\in\mathbb{R}^{2\times d_h} \\
\mathbf{V}_h &=
\begin{bmatrix}
W_h^{V,I}\mathbf{f}_I \\
W_h^{V,M}\mathbf{f}_M
\end{bmatrix}
\in\mathbb{R}^{2\times d_h}
\end{aligned}
\end{equation}

where $W_h^{Q,I}$, $W_h^{K,I}$, $W_h^{V,I}$ $\!\in\!\mathbb{R}^{d_h\times D_I}$ and
$W_h^{Q,M}$, $W_h^{K,M}$, $W_h^{V,M}$ $\!\in\!\mathbb{R}^{d_h\times D_M}$.

\paragraph{Two-token multi-head attention.}
Each head computes a \(2\times 2\) attention over the two modalities and mixes the values
\begin{gather}
\mathbf{U}_h
= \mathrm{softmax}\!\left(\frac{\mathbf{Q}_h \mathbf{K}_h^{\top}}{\sqrt{d_h}}\right)\mathbf{V}_h
\in \mathbb{R}^{2\times d_h}
\\
\mathbf{o}
= W^{O}\mathrm{Concat}(\mathrm{vec}(\mathbf{U}_1),\ldots,\mathrm{vec}(\mathbf{U}_H))
\in \mathbb{R}^{D_{IM}}
\end{gather}
with \(W^{O}\in\mathbb{R}^{D_{IM}\times(2H d_h)}\). Here \(\mathrm{vec}(\cdot)\) stacks row-wise and \(\mathrm{Concat}(\cdot)\) concatenates vectors.

\paragraph{Output and residual path.}
The attention output is re-projected and merged with the raw features via a residual connection
\begin{equation}
\mathbf{f}_{IM}
= W_{\mathrm{skip}}\,[\,\mathbf{f}_I \,\|\, \mathbf{f}_M\,]
\;+\; g(\mathbf{o})
\;\in\; \mathbb{R}^{D_{IM}}
\end{equation}
where \(W_{\mathrm{skip}}\!\in\!\mathbb{R}^{D_{IM}\times(D_I+D_M)}\), \(g(\cdot)\) is a linear layer followed by a pointwise nonlinearity, and \([\,\cdot \| \cdot\,]\) denotes vector concatenation.
This design explicitly models self and mutual interactions (through the \(2\times 2\) attention) while keeping a skip path that preserves modality-specific cues and maintains stable gradients back to \(M_I\) and \(M_M\).
\section{Results}

% \begin{table*}[ht]
% \centering
% \small
% \setlength{\tabcolsep}{4pt}
% \begin{tabular}{@{}p{0.3\linewidth}cccccc@{}}
% \toprule
% \textbf{Method} &
% \textbf{AUC} &
% \textbf{Precision} &
% \textbf{Accuracy} &
% \textbf{Sensitivity} &
% \textbf{Specificity} &
% \textbf{Dice} \\
% \midrule
% JIF-MMFA \cite{TANG2024110604} & 0.750 & 0.328 & 0.896 & 0.406 & 0.939 & 0.306 \\
% VEMFL \cite{restrepo2024multimodaldeeplearninglowresource} & 0.842 & 0.439 & 0.912 & 0.291 & 0.958 & 0.302 \\
% CAFFM \cite{TRANVAN2025102588} & 0.787 & 0.306 & 0.898 & 0.233 & 0.954 & 0.232 \\
% ALBEF \cite{Adebiyi2024.05.30.24308213} & 0.760 & 0.294 & 0.889 & 0.208 & 0.945 & 0.222 \\
% CoscatNet-UFS \cite{ZUO2025103091} & 0.800 & 0.472 & 0.925 & 0.453 & \textbf{0.959} & 0.435 \\
% DualRefNet \cite{Khurshid2025} & 0.848 & 0.447 & 0.918 & 0.422 & 0.954 & 0.441 \\
% SkinM2Former \cite{10943957} & 0.841 & 0.436 & 0.920 & 0.421 & 0.954 & 0.405 \\
% Fine-tuned PanDerm \cite{Yan2025} & 0.840 & 0.476 & 0.920 & 0.447 & 0.953 & 0.434 \\
% \midrule
% \textbf{JI-ADF (ours)} & \textbf{0.866} & \textbf{0.543} & \textbf{0.930} & \textbf{0.536} & \textbf{0.959} & \textbf{0.505} \\
% \bottomrule
% \end{tabular}
% \caption{Comparison with state-of-the-art skin lesion classification approaches.}
% \label{tab:comparision}
% \end{table*}

Across competitive multimodal baselines (Table~\ref{tab:comparision}), JI-ADF consistently achieves the best overall performance, with the highest AUC (0.866) and accuracy (0.930), indicating strong class separability and stable predictions despite the heterogeneity and long-tailed nature of the MILK10k dataset. Sensitivity remains particularly challenging under severe class imbalance; nevertheless, JI-ADF attains the top sensitivity (0.536), suggesting improved true-positive detection, especially in ambiguous cases where visual cues alone may be insufficient. This gain is achieved without sacrificing precision, as reflected in the highest Dice (0.505) and leading precision (0.543), indicating a better balance between recall and false positives. In addition, JI-ADF matches or slightly exceeds the best specificity (0.959), further demonstrating well-calibrated predictions. Overall, these results suggest that adaptive, per-sample fusion of multimodal inputs provides a more balanced and reliable decision strategy than fixed fusion approaches.

\section{Ablation Study}

\subsection{Modality Contribution Analysis}

\begin{table}[h]
\centering
\small
\setlength{\tabcolsep}{1.2pt}
\begin{tabular}{@{}p{0.13\linewidth}cccccc@{}}
\toprule
\textbf{Config.} &
\textbf{AUC} &
\textbf{Precision} &
\textbf{Accuracy} &
\textbf{Sensitivity} &
\textbf{Specificity} &
\textbf{Dice} \\
\midrule
C  & 0.799 & 0.415 & 0.919 & 0.391 & 0.952 & 0.392 \\
D  & 0.807 & 0.458 & 0.923 & 0.416 & 0.954 & 0.420 \\
M  & 0.796 & 0.334 & 0.894 & 0.367 & 0.938 & 0.325 \\
C+D & \textbf{0.866} & 0.487 & 0.925 & 0.464 & 0.955 & 0.456 \\
C+M & 0.834 & 0.467 & 0.918 & 0.389 & 0.952 & 0.376 \\
D+M & 0.829 & 0.446 & 0.918 & 0.451 & 0.952 & 0.413 \\
\textbf{C+D+M} & \textbf{0.866} & \textbf{0.543} & \textbf{0.930} & \textbf{0.536} & \textbf{0.959} & \textbf{0.505} \\
\bottomrule
\end{tabular}
\caption{Modality Configuration Ablation Study. (C: clinical image, D: dermoscopic image, M: metadata).}
\label{tab:modalconfig}
\end{table}

We conducted an ablation study over all unimodal and bimodal subsets of the inputs. Based on the results shown in Table~\ref{tab:modalconfig}, models that combine modalities consistently outperform single-modality variants. In particular, the proposed JI-ADF trimodal fusion achieves the best performance across most metrics, indicating that jointly leveraging clinical images, dermoscopic images, and metadata yields the most reliable classifier.

\subsection{Ablation of Attention and Residual Branches}

The attention and residual paths in MMFA serve complementary roles. The attention path explicitly models cross-modal dependencies, enabling metadata to modulate image features when strong semantic correlations exist. In contrast, the residual path preserves modality-specific evidence by allowing the original image and metadata features to directly contribute to the fused representation. When using attention alone, the fused representation relies entirely on learned cross-modal correlations, which can be unstable under weak or noisy inter-modal relationships. Conversely, the skip-only variant lacks the capacity to capture fine-grained interactions across modalities. Combining both paths allows the model to dynamically balance cross-modal reasoning and modality-specific evidence, leading to more stable and expressive fusion. The superior Sensitivity and Dice achieved by the full MMFA confirm that the residual branch is not merely an optimization shortcut, but a structural component that complements attention by safeguarding reliable unimodal signals.

\begin{table}[!htbp]
\centering
\small
\setlength{\tabcolsep}{4pt}
\resizebox{\columnwidth}{!}{%
\begin{tabular}{lcccccc}
\toprule
\textbf{Fusion Variant} &
\textbf{AUC} &
\textbf{Precision} &
\textbf{Accuracy} &
\textbf{Sensitivity} &
\textbf{Specificity} &
\textbf{Dice} \\
\midrule
Skip-only         & 0.880 & 0.508 & 0.927 & 0.421 & 0.960 & 0.419 \\
Attention-only    & 0.882 & 0.542 & 0.930 & 0.438 & 0.963 & 0.457 \\
Attention + Skip  & 0.866 & 0.543 & 0.930 & 0.536 & 0.959 & 0.505 \\
\bottomrule
\end{tabular}
}
\caption{Ablation of attention and residual branches in the Multimodal Fusion Attention (MMFA) module.}
% \caption*{Skip-only corresponds to residual fusion without attention, while Attention-only removes the residual path and relies solely on cross-attention.}
\label{tab:mmfaconfig}
\end{table}

\subsection{Fusion Mechanism Ablation}

To examine the effect of different fusion strategies, we compare a sequence of architectural variants that progressively increase the capacity for cross-modal interaction. \textbf{Late concat} simply merges the two image embeddings and metadata at the final classifier. \textbf{JF-concat} retains this linear merging but introduces three prediction heads trained with auxiliary losses. \textbf{JF-MMFA} replaces concatenation with a multimodal attention block to produce a unified representation, using a single joint head for prediction. \textbf{JI-MMFA} reinstates the three-head design on top of the attention module and combines their outputs through fixed averaging. \textbf{JI-ADF (no aux)} preserves the three-head structure but substitutes fixed averaging with a learnable adaptive fusion module that assigns instance-dependent weights.

% \begin{table}[ht]
% \centering
% \caption{\textbf{Fusion Architecture Ablation Study.}}
% \label{tab:comparision1}
% \begingroup
% \small % chữ nhỏ hơn
% \setlength{\tabcolsep}{4pt} % giảm khoảng cách giữa các cột
% \begin{tabular}{@{}p{0.20\linewidth}cccccc@{}} % tăng cột đầu lên 22% linewidth
% \toprule
% \textbf{Fusion Method} &
% \textbf{AUC} &
% \textbf{Precision} &
% \textbf{Accuracy} &
% \textbf{Sensitivity} &
% \textbf{Specificity} &
% \textbf{Dice} \\
% \midrule
% % VEMFL (Vector Embedding Multimodal Fusion Learning)
% Late concat & 0.893 & 0.517 & 0.927 & 0.337 & \textbf{0.967} & 0.360 \\
% JF-concat & 0.858 & 0.484 & 0.921 & 0.416 & 0.957 & 0.419 \\
% JF-MMFA & 0.887 & 0.544 & \textbf{0.930} & 0.420 & 0.963 & 0.439 \\
% JI-MMFA & 0.895 & \textbf{0.555} & \textbf{0.930} & 0.453 & 0.963 & 0.462 \\
% JI-ADF (no aux) & \textbf{0.899} & 0.551 & 0.927 & 0.477 & 0.961 & 0.489 \\

% \textbf{JI-ADF (ours)} & 0.866 & 0.543 & \textbf{0.930} & \textbf{0.536} & 0.959 & \textbf{0.505} \\
% \bottomrule
% \end{tabular}
% \endgroup
% \end{table}

\begin{figure}[!ht]
\centering
\includegraphics[width=\linewidth]{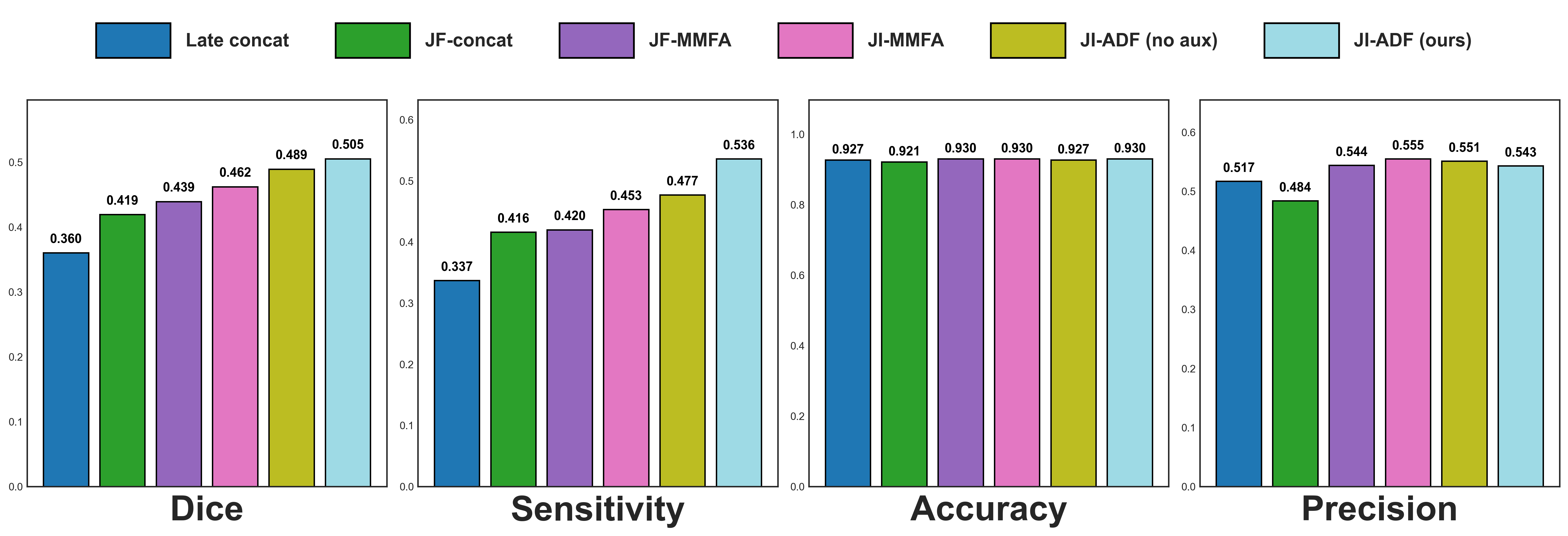}
\caption{Fusion Architecture Ablation – Multimetrics Comparison}
\label{fig:fusion_ablation}
\end{figure}

Across the four metrics reported in Figure~\ref{fig:fusion_ablation}, the ablation results show a steady improvement as the fusion design becomes more expressive. MMFA strengthens joint feature learning, with JF-MMFA outperforming both Late concat and JF-concat, suggesting that attention-based fusion captures complementary information more effectively than simple merging. Adding individual heads with auxiliary supervision further stabilizes the optimization process, enabling JI-MMFA to surpass the single-head JF-MMFA. Replacing fixed averaging with adaptive weighting introduces another consistent step forward, as JI-ADF (no aux) benefits from instance-dependent fusion that adjusts to the reliability of each modality. Bringing these components together in the full JI-ADF model yields the most balanced and robust performance overall, indicating that adaptive fusion and auxiliary supervision work together to produce a more stable, balanced, and reliably integrated multimodal representation.
\section{Conclusions}
In this work, we presented JI-ADF, a trimodal architecture for the multi-class classification of skin lesions. By introducing a unified attention-based fusion block, our model captures cross-modal interactions between dermoscopic images, clinical images, and structured metadata. Combined with class-aware optimization, JI-ADF improves diagnostic performance across both common and underrepresented conditions. Evaluations on the MILK10k benchmark confirm its strong generalization and robustness. We hope that the design principles of JI-ADF offer a scalable foundation for future diagnostic frameworks, particularly in diseases requiring multimodal diagnosis. For example, diagnosing complex conditions like endometriosis often involves identifying multiple lesion sites across different imaging sources, such as MRI and ultrasound.

\newpage


\bibliography{example_paper}

@article{Yan2025,
  author  = {Yan, Siyuan and Yu, Zhen and Primiero, Clare and Vico-Alonso, Cristina and Wang, Zhonghua and Yang, Litao and Tschandl, Philipp and Hu, Ming and Ju, Lie and Tan, Gin and Tang, Vincent and Ng, Aik Beng and Powell, David and Bonnington, Paul and See, Simon and Magnaterra, Elisabetta and Ferguson, Peter and Nguyen, Jennifer and Guitera, Pascale and Banuls, Jose and Janda, Monika and Mar, Victoria and Kittler, Harald and Soyer, H. Peter and Ge, Zongyuan},
  title   = {A multimodal vision foundation model for clinical dermatology},
  journal = {Nature Medicine},
  year    = {2025},
  volume  = {31},
  number  = {8},
  pages   = {2691--2702},
  month   = aug,
  doi     = {10.1038/s41591-025-03747-y},
  url     = {https://doi.org/10.1038/s41591-025-03747-y}
}

@article{bray2024globalcancer2022,
  author  = {Bray, Freddie and Laversanne, Mathieu and Sung, Hyuna and Ferlay, Jacques and Siegel, Rebecca L. and Soerjomataram, Isabelle and Jemal, Ahmedin},
  title   = {Global cancer statistics 2022: {GLOBOCAN} estimates of incidence and mortality worldwide for 36 cancers in 185 countries},
  journal = {CA: A Cancer Journal for Clinicians},
  year    = {2024},
  volume  = {74},
  number  = {3},
  pages   = {229--263},
  doi     = {10.3322/caac.21834},
  pmid    = {38572751}
}

@misc{acs_cancerfacts2024,
  author       = {{American Cancer Society}},
  title        = {Cancer Facts \& Figures 2024},
  year         = {2024},
  address      = {Atlanta},
  institution  = {American Cancer Society},
  url = {https://www.cancer.org/content/dam/cancer-org/research/cancer-facts-and-statistics/annual-cancer-facts-and-figures/2024/2024-cancer-facts-and-figures-acs.pdf},
  
}

@misc{ferlay2024cancertomorrow,
  author      = {Ferlay, J. and Laversanne, M. and Ervik, M. and Lam, F. and Colombet, M. and Mery, L. and Pi{\~n}eros, M. and Znaor, A. and Soerjomataram, I. and Bray, F.},
  title       = {Global Cancer Observatory: Cancer Tomorrow (version 1.1)},
  year        = {2024},
  institution = {International Agency for Research on Cancer},
  address     = {Lyon, France},
  url = {https://gco.iarc.who.int/tomorrow},
}

@article{Tschandl2025MILK10k,
  title   = {MILK10k: A Hierarchical Multimodal Imaging-Learning Toolkit for Diagnosing Pigmented and Nonpigmented Skin Cancer and its Simulators},
  author  = {Tschandl, Philipp and Akay, Bengu Nisa and Rosendahl, Cliff and Rotemberg, Veronica and Todorovska, Verche and Weber, Jochen and Wolber, Anna Katharina and M{\"u}ller, Christoph and Kurtansky, Nicholas and Halpern, Allan and Weninger, Wolfgang and Kittler, Harald},
  journal = {Journal of Investigative Dermatology},
  year    = {2025},
  doi     = {10.1016/j.jid.2025.06.1594},
  issn    = {0022-202X},
  publisher = {Elsevier},
  url     = {https://doi.org/10.1016/j.jid.2025.06.1594}
}

@article{TANG2024110604,
title = {Joint-individual fusion structure with fusion attention module for multi-modal skin cancer classification},
journal = {Pattern Recognition},
volume = {154},
pages = {110604},
year = {2024},
issn = {0031-3203},
doi = {https://doi.org/10.1016/j.patcog.2024.110604},
url = {https://www.sciencedirect.com/science/article/pii/S0031320324003558},
author = {Peng Tang and Xintong Yan and Yang Nan and Xiaobin Hu and Bjoern H. Menze and Sebastian Krammer and Tobias Lasser}
}

@article{TRANVAN2025102588,
title = {A multimodal skin lesion classification through cross-attention fusion and collaborative edge computing},
journal = {Computerized Medical Imaging and Graphics},
volume = {124},
pages = {102588},
year = {2025},
issn = {0895-6111},
doi = {https://doi.org/10.1016/j.compmedimag.2025.102588},
url = {https://www.sciencedirect.com/science/article/pii/S0895611125000977},
author = {Nhu-Y Tran-Van and Kim-Hung Le}
}

@article {Adebiyi2024.05.30.24308213,
	author = {Adebiyi, Abdulmateen and Abdalnabi, Nader and Smith, Emily Hoffman and Hirner, Jesse and Simoes, Eduardo J. and Becevic, Mirna and Rao, Praveen},
	title = {Accurate Skin Lesion Classification Using Multimodal Learning on the HAM10000 and ISIC 2017 Datasets},
	elocation-id = {2024.05.30.24308213},
	year = {2025},
	doi = {10.1101/2024.05.30.24308213},
	publisher = {Cold Spring Harbor Laboratory Press},
	URL = {https://www.medrxiv.org/content/early/2025/05/20/2024.05.30.24308213},
	eprint = {https://www.medrxiv.org/content/early/2025/05/20/2024.05.30.24308213.full.pdf},
	journal = {medRxiv}
}

@misc{restrepo2024multimodaldeeplearninglowresource,
      title={Multimodal Deep Learning for Low-Resource Settings: A Vector Embedding Alignment Approach for Healthcare Applications}, 
      author={David Restrepo and Chenwei Wu and Sebastián Andrés Cajas and Luis Filipe Nakayama and Leo Anthony Celi and Diego M López},
      year={2024},
      eprint={2406.02601},
      archivePrefix={arXiv},
      primaryClass={cs.LG},
      url={https://arxiv.org/abs/2406.02601}, 
}

@article{ZUO2025103091,
title = {A multi-stage multi-modal learning algorithm with adaptive multimodal fusion for improving multi-label skin lesion classification},
journal = {Artificial Intelligence in Medicine},
volume = {162},
pages = {103091},
year = {2025},
issn = {0933-3657},
doi = {https://doi.org/10.1016/j.artmed.2025.103091},
url = {https://www.sciencedirect.com/science/article/pii/S0933365725000260},
author = {Lihan Zuo and Zizhou Wang and Yan Wang}
}

@article{Khurshid2025,
  author    = {Mahapara Khurshid and Richa Singh and Mayank Vatsa},
  title     = {Multimodal dual-stage feature refinement for robust skin lesion classification},
  journal   = {Scientific Reports},
  year      = {2025},
  volume    = {15},
  number    = {1},
  pages     = {37775},
  doi       = {10.1038/s41598-025-14839-7},
  url       = {https://doi.org/10.1038/s41598-025-14839-7},
  issn      = {2045-2322}
}

@INPROCEEDINGS {10943957,
author = { Zhang, Yuan and Xie, Yutong and Wang, Hu and Avery, Jodie C and Hull, M Louise and Carneiro, Gustavo },
booktitle = { 2025 IEEE/CVF Winter Conference on Applications of Computer Vision (WACV) },
title = {{ A Novel Perspective for Multi-Modal Multi-Label Skin Lesion Classification }},
year = {2025},
volume = {},
ISSN = {},
pages = {3549-3558},
doi = {10.1109/WACV61041.2025.00350},
url = {https://doi.ieeecomputersociety.org/10.1109/WACV61041.2025.00350},
publisher = {IEEE Computer Society},
address = {Los Alamitos, CA, USA},
month =mar}

@misc{loshchilov2019decoupledweightdecayregularization,
      title={Decoupled Weight Decay Regularization}, 
      author={Ilya Loshchilov and Frank Hutter},
      year={2019},
      eprint={1711.05101},
      archivePrefix={arXiv},
      primaryClass={cs.LG},
      url={https://arxiv.org/abs/1711.05101}, 
}

@INPROCEEDINGS{5206848,
  author={Deng, Jia and Dong, Wei and Socher, Richard and Li, Li-Jia and Kai Li and Li Fei-Fei},
  booktitle={2009 IEEE Conference on Computer Vision and Pattern Recognition}, 
  title={ImageNet: A large-scale hierarchical image database}, 
  year={2009},
  volume={},
  number={},
  pages={248-255},
  doi={10.1109/CVPR.2009.5206848}
}

@ARTICLE{9018274,
  author={Tang, Peng and Liang, Qiaokang and Yan, Xintong and Xiang, Shao and Zhang, Dan},
  journal={IEEE Journal of Biomedical and Health Informatics}, 
  title={GP-CNN-DTEL: Global-Part CNN Model With Data-Transformed Ensemble Learning for Skin Lesion Classification}, 
  year={2020},
  volume={24},
  number={10},
  pages={2870-2882},
  doi={10.1109/JBHI.2020.2977013}}

@ARTICLE{8990108,
  author={Xie, Yutong and Zhang, Jianpeng and Xia, Yong and Shen, Chunhua},
  journal={IEEE Transactions on Medical Imaging}, 
  title={A Mutual Bootstrapping Model for Automated Skin Lesion Segmentation and Classification}, 
  year={2020},
  volume={39},
  number={7},
  pages={2482-2493},
  doi={10.1109/TMI.2020.2972964}}

@ARTICLE{7792699,
  author={Yu, Lequan and Chen, Hao and Dou, Qi and Qin, Jing and Heng, Pheng-Ann},
  journal={IEEE Transactions on Medical Imaging}, 
  title={Automated Melanoma Recognition in Dermoscopy Images via Very Deep Residual Networks}, 
  year={2017},
  volume={36},
  number={4},
  pages={994-1004},
  doi={10.1109/TMI.2016.2642839}}

@article{BI2020107502,
title = {Multi-Label classification of multi-modality skin lesion via hyper-connected convolutional neural network},
journal = {Pattern Recognition},
volume = {107},
pages = {107502},
year = {2020},
issn = {0031-3203},
doi = {https://doi.org/10.1016/j.patcog.2020.107502},
url = {https://www.sciencedirect.com/science/article/pii/S0031320320303058},
author = {Lei Bi and David Dagan Feng and Michael Fulham and Jinman Kim}
}

@article{Esteva2017,
  author  = {Esteva, Andre and Kuprel, Brett and Novoa, Roberto A. and Ko, Justin and Swetter, Susan M. and Blau, Helen M. and Thrun, Sebastian},
  title   = {Dermatologist-level classification of skin cancer with deep neural networks},
  journal = {Nature},
  year    = {2017},
  volume  = {542},
  number  = {7639},
  pages   = {115--118},
  doi     = {10.1038/nature21056},
  url     = {https://doi.org/10.1038/nature21056},
  issn    = {1476-4687}
}

@article{Haenssle2018,
  author  = {Haenssle, H. A. and Fink, C. and Schneiderbauer, R. and 
             Toberer, F. and Buhl, T. and Blum, A. and Kalloo, A. and 
             Hassen, A. Ben Hadj and Thomas, L. and Enk, A. and Uhlmann, L. 
             and Reader Study Level-I and Level-II Groups},
  title   = {Man against machine: diagnostic performance of a deep learning convolutional neural network for dermoscopic melanoma recognition in comparison to 58 dermatologists},
  journal = {Annals of Oncology},
  year    = {2018},
  volume  = {29},
  number  = {8},
  pages   = {1836--1842},
  month   = {Aug},
  doi     = {10.1093/annonc/mdy166},
  url     = {https://doi.org/10.1093/annonc/mdy166},
  issn    = {0923-7534},
  pmid    = {29846502},
  publisher = {Oxford University Press}
}

@article{Atrey2010,
  author  = {Atrey, Pradeep K. and Hossain, M. Anwar and El Saddik, Abdulmotaleb and Kankanhalli, Mohan S.},
  title   = {Multimodal fusion for multimedia analysis: a survey},
  journal = {Multimedia Systems},
  year    = {2010},
  volume  = {16},
  number  = {6},
  pages   = {345--379},
  month   = {Nov},
  doi     = {10.1007/s00530-010-0182-0},
  url     = {https://doi.org/10.1007/s00530-010-0182-0},
  issn    = {1432-1882}
}

@article{Huang2020,
  author  = {Huang, Shih-Cheng and Pareek, Anuj and Seyyedi, Saeed and Banerjee, Imon and Lungren, Matthew P.},
  title   = {Fusion of medical imaging and electronic health records using deep learning: a systematic review and implementation guidelines},
  journal = {npj Digital Medicine},
  year    = {2020},
  volume  = {3},
  number  = {1},
  pages   = {136},
  doi     = {10.1038/s41746-020-00341-z},
  url     = {https://doi.org/10.1038/s41746-020-00341-z},
  issn    = {2398-6352}
}

@ARTICLE{9364366,
  author={Pacheco, Andre G. C. and Krohling, Renato A.},
  journal={IEEE Journal of Biomedical and Health Informatics}, 
  title={An Attention-Based Mechanism to Combine Images and Metadata in Deep Learning Models Applied to Skin Cancer Classification}, 
  year={2021},
  volume={25},
  number={9},
  pages={3554-3563},
  doi={10.1109/JBHI.2021.3062002}}

@article{Yap2018,
  author  = {Yap, Jordan and Yolland, William and Tschandl, Philipp},
  title   = {Multimodal skin lesion classification using deep learning},
  journal = {Experimental Dermatology},
  year    = {2018},
  volume  = {27},
  number  = {11},
  pages   = {1261--1267},
  month   = {Nov},
  doi     = {10.1111/exd.13777},
  issn    = {0906-6705},
  pmid    = {30187575},
  publisher = {John Wiley \& Sons Ltd}
}

@article{Liu2020,
  author  = {Liu, Yuan and Jain, Ayush and Eng, Clara and Way, David H. and Lee, Kang and 
             Bui, Peggy and Kanada, Kimberly and de Oliveira Marinho, Guilherme and 
             Gallegos, Jessica and Gabriele, Sara and Gupta, Vishakha and Singh, Nalini and 
             Natarajan, Vivek and Hofmann-Wellenhof, Rainer and Corrado, Greg S. and 
             Peng, Lily H. and Webster, Dale R. and Ai, Dennis and Huang, Susan J. and 
             Liu, Yun and Dunn, R. Carter and Coz, David},
  title   = {A deep learning system for differential diagnosis of skin diseases},
  journal = {Nature Medicine},
  year    = {2020},
  volume  = {26},
  number  = {6},
  pages   = {900--908},
  month   = {Jun},
  doi     = {10.1038/s41591-020-0842-3},
  url     = {https://doi.org/10.1038/s41591-020-0842-3},
  issn    = {1546-170X}
}

@INPROCEEDINGS{9098645,
  author={Li, Weipeng and Zhuang, Jiaxin and Wang, Ruixuan and Zhang, Jianguo and Zheng, Wei-Shi},
  booktitle={2020 IEEE 17th International Symposium on Biomedical Imaging (ISBI)}, 
  title={Fusing Metadata and Dermoscopy Images for Skin Disease Diagnosis}, 
  year={2020},
  volume={},
  number={},
  pages={1996-2000},
  doi={10.1109/ISBI45749.2020.9098645}}

@article{PACHECO2020103545,
title = {The impact of patient clinical information on automated skin cancer detection},
journal = {Computers in Biology and Medicine},
volume = {116},
pages = {103545},
year = {2020},
issn = {0010-4825},
doi = {https://doi.org/10.1016/j.compbiomed.2019.103545},
url = {https://www.sciencedirect.com/science/article/pii/S0010482519304019},
author = {Andre G.C. Pacheco and Renato A. Krohling}
}

@article{Cai2023,
  author  = {Cai, Gan and Zhu, Yu and Wu, Yue and Jiang, Xiaoben and Ye, Jiongyao and Yang, Dawei},
  title   = {A multimodal transformer to fuse images and metadata for skin disease classification},
  journal = {The Visual Computer},
  year    = {2023},
  volume  = {39},
  number  = {7},
  pages   = {2781--2793},
  month   = {Jul},
  doi     = {10.1007/s00371-022-02492-4},
  url     = {https://doi.org/10.1007/s00371-022-02492-4},
  issn    = {1432-2315}
}

@ARTICLE{9363019,
  author={He, Xingxin and Deng, Ying and Fang, Leyuan and Peng, Qinghua},
  journal={IEEE Transactions on Medical Imaging}, 
  title={Multi-Modal Retinal Image Classification With Modality-Specific Attention Network}, 
  year={2021},
  volume={40},
  number={6},
  pages={1591-1602},
  doi={10.1109/TMI.2021.3059956}}

@ARTICLE{8027090,
  author={Hu, Junlin and Lu, Jiwen and Tan, Yap-Peng},
  journal={IEEE Transactions on Pattern Analysis and Machine Intelligence}, 
  title={Sharable and Individual Multi-View Metric Learning}, 
  year={2018},
  volume={40},
  number={9},
  pages={2281-2288},
  doi={10.1109/TPAMI.2017.2749576}}
\bibliographystyle{icml2026}

%%%%%%%%%%%%%%%%%%%%%%%%%%%%%%%%%%%%%%%%%%%%%%%%%%%%%%%%%%%%%%%%%%%%%%%%%%%%%%%
%%%%%%%%%%%%%%%%%%%%%%%%%%%%%%%%%%%%%%%%%%%%%%%%%%%%%%%%%%%%%%%%%%%%%%%%%%%%%%%
% APPENDIX
%%%%%%%%%%%%%%%%%%%%%%%%%%%%%%%%%%%%%%%%%%%%%%%%%%%%%%%%%%%%%%%%%%%%%%%%%%%%%%%
%%%%%%%%%%%%%%%%%%%%%%%%%%%%%%%%%%%%%%%%%%%%%%%%%%%%%%%%%%%%%%%%%%%%%%%%%%%%%%%
\newpage
\appendix
\onecolumn
\section{Experiments}

\subsection{Dataset Description} 
We used the MILK10k multimodal skin-lesion dataset \cite{Tschandl2025MILK10k} to train and evaluate our method. The MILK10k dataset contains 10480 images from 5240 lesions, provided as paired clinical close-up and dermoscopic images collected across five centers. For testing, we follow the official hidden test comprising 479 lesions with paired images (958 images), provided with the metadata fields as the training set. This data can be found here: \url{https://challenge.isic-archive.com/data/#milk10k}.

\begin{table}[ht]
\centering
\small
\renewcommand{\arraystretch}{0.9}
\setlength{\tabcolsep}{8pt}
\resizebox{0.9\columnwidth}{!}{%
\begin{tabular}{lcc}
\toprule
\textbf{Diagnostic Category} & \textbf{Abbreviation} & \textbf{Quantity (training set)} \\
\midrule
Actinic keratosis/intraepidermal carcinoma & AKIEC & 242 \\
Basal cell carcinoma & BCC & 2017 \\
Other benign proliferations including collisions & BEN\_OTH & 35 \\
Benign keratinocytic lesion & BKL & 435 \\
Dermatofibroma & DF & 42 \\
Inflammatory and infectious & INF & 40 \\
Other malignant proliferations including collisions & MAL\_OTH & 7 \\
Melanoma & MEL & 360 \\
Melanocytic nevus, any type & NV & 597 \\
Squamous cell carcinoma/keratoacanthoma & SCCKA & 379 \\
Vascular lesions and hemorrhage & VASC & 38 \\
\bottomrule
\end{tabular}
}
\caption{Distribution of diagnostic categories in the training set.}
\label{tab:data_distribution}
\end{table}

\subsection{Implementation}

% \cite{loshchilov2019decoupledweightdecayregularization}

Our model was trained for 50 epochs with a batch size of $16$, using AdamW optimizer \cite{loshchilov2019decoupledweightdecayregularization} with the initial learning rate value of 1e-4 and weight decay of 1e-5. The ReduceLROnPlateau scheduler is applied for the learning rate decay. We split the training set into 80\% for training and 20\% for validation. The model achieving the highest average Macro F1 Score on the validation set was saved for testing. The ImageNet-1K \cite{5206848} pretrained EfficientNetV2 is employed as the backbone. Input images are resized to $384 \times 384 \times 3$, the length of the encoded patient's metadata is $256$, and we use MMFA (Section~\ref{sec:MMFA}) to fuse images and metadata features. All the experiments were performed using Python 3.12 with PyTorch 2.8.0 and run on NVIDIA A100 GPU with 40GB VRAM.

\section{Joint Fusion Structure}

For convenience, we consider the fusion of a dermoscopic image, a clinical close-up image, and patient metadata for skin lesion diagnosis as a multi-class classification task. Each case contains a dermoscopic image $I_{\mathrm{derm}}$, a close-up image $I_{\mathrm{close}}$, patient metadata $M$, and a label $y\in\{1,\dots,N\}$.

$M_I$ is the model to extract features of 2 images and $M_M$ is the method to extract features from patient metadata $M$:
\begin{equation}
\begin{gathered}
\mathbf{f}_I = M_I(I_{\mathrm{derm}}, I_{\mathrm{close}}) \in \mathbb{R}^{D_I} \\
\mathbf{f}_M = M_M(M) \in \mathbb{R}^{D_M}
\end{gathered}
\end{equation}

A fusion module $FM$ produces the joint representation
\begin{equation}
\mathbf{f}_{IM} = FM(\mathbf{f}_I, \mathbf{f}_M) \in \mathbb{R}^{D_{IM}}
\end{equation}
which a classifier $C_{IM}$ maps to logits $\mathbf{z}_{IM} = C_{IM}(\mathbf{f}_{IM})$ and posteriors
\begin{equation}
\mathbf{P}_{IM} = \mathrm{softmax}(\mathbf{z}_{IM}) \in \Delta^{N-1}
\end{equation}

With one-hot target $\mathbf{y}$, the joint loss used in the figure is
\begin{equation}
\mathcal{L}_{IM} = -\sum_{c=1}^{N} y_c \log P_{IM}^{(c)}
\end{equation}

Because $\mathbf{f}_{IM}$ depends on both streams, the loss backpropagates through the fusion node:
\begin{align}
\frac{\partial \mathcal{L}_{IM}}{\partial \theta_{M_I}}
&=
\frac{\partial \mathcal{L}_{IM}}{\partial \mathbf{z}_{IM}}
\frac{\partial \mathbf{z}_{IM}}{\partial \mathbf{f}_{IM}}
\frac{\partial \mathbf{f}_{IM}}{\partial \mathbf{f}_I}
\frac{\partial \mathbf{f}_I}{\partial \theta_{M_I}} \\
\frac{\partial \mathcal{L}_{IM}}{\partial \theta_{M_M}}
&=
\frac{\partial \mathcal{L}_{IM}}{\partial \mathbf{z}_{IM}}
\frac{\partial \mathbf{z}_{IM}}{\partial \mathbf{f}_{IM}}
\frac{\partial \mathbf{f}_{IM}}{\partial \mathbf{f}_M}
\frac{\partial \mathbf{f}_M}{\partial \theta_{M_M}}
\end{align}

\textbf{Summary.} Inputs $(I_{\mathrm{derm}}, I_{\mathrm{close}}, M)$ are encoded into $(\mathbf{f}_I,\mathbf{f}_M)$, fused by $FM$ into $\mathbf{f}_{IM}$, and classified by $C_{IM}$ to yield $\mathbf{P}_{IM}$, matching the forward and gradient flows in Figure~\ref{fig:2_methods}.

\section{Performance of the proposed model}

\begin{table*}[ht]
\centering
\scriptsize
\setlength{\tabcolsep}{6pt}
\begin{tabular}{l c c c c c c c c c c c c}
\toprule
{\textbf{Category Metric}} & {\textbf{Mean}} & \multicolumn{11}{c}{\textbf{Diagnosis Category}} \\
\cmidrule(lr){3-13}
 & & AKIEC & BCC & BEN\_OTH & BKL & DF & INF & MAL\_OTH & MEL & NV & SCCKA & VASC \\
\midrule
AUC                & 0.866 & 0.903 & 0.905 & 0.837 & 0.834 & 1.000 & 0.757 & 0.521 & 0.933 & 0.932 & 0.912 & 0.992 \\
AUC, Sens $>$ 80\% & 0.734 & 0.816 & 0.832 & 0.540 & 0.632 & 1.000 & 0.449 & 0.214 & 0.895 & 0.862 & 0.848 & 0.985 \\
Average Precision  & 0.543 & 0.690 & 0.543 & 0.141 & 0.591 & 1.000 & 0.113 & 0.026 & 0.688 & 0.790 & 0.769 & 0.625 \\
Accuracy           & 0.930 & 0.881 & 0.835 & 0.983 & 0.850 & 1.000 & 0.956 & 0.979 & 0.950 & 0.946 & 0.866 & 0.985 \\
Sensitivity        & 0.536 & 0.592 & 0.902 & 0.000 & 0.440 & 1.000 & 0.273 & 0.000 & 0.769 & 0.654 & 0.667 & 0.600 \\
Specificity        & 0.959 & 0.931 & 0.825 & 0.998 & 0.958 & 1.000 & 0.972 & 1.000 & 0.966 & 0.981 & 0.931 & 0.989 \\
Dice Coefficient   & 0.505 & 0.596 & 0.582 & 0.000 & 0.550 & 1.000 & 0.222 & 0.000 & 0.714 & 0.723 & 0.709 & 0.462 \\
PPV                & 0.596 & 0.600 & 0.430 & 0.000 & 0.733 & 1.000 & 0.188 & 1.000 & 0.667 & 0.810 & 0.757 & 0.375 \\
NPV                & 0.960 & 0.929 & 0.983 & 0.985 & 0.866 & 1.000 & 0.983 & 0.979 & 0.979 & 0.959 & 0.896 & 0.996 \\
\bottomrule
\end{tabular}
\caption{Performance of the proposed JI-ADF method across all diagnostic categories. Each column reports per-class results for the evaluation metrics. The Mean Value column represents the macro-averaged score across all 11 lesion types.}
\label{tab:lesion-metrics}
\end{table*}

The proposed JI-ADF model delivers strong and balanced performance across categories. As shown in Table~\ref{tab:lesion-metrics}, it reaches a mean AUC of 0.866, overall accuracy of 0.930, specificity of 0.959, and NPV of 0.960, indicating reliable discrimination with low false-negative rates. The moderate mean sensitivity (0.536) and Dice score (0.505) align with the severe class imbalance in the dataset (Table~\ref{tab:data_distribution}). The model excels in abundant and visually distinctive classes such as BCC, NV, and DF, achieving high AUCs (0.905–1.000) and accuracies above 0.93, suggesting effective use of both image and metadata cues. In contrast, classes with very limited samples or high variability (BEN\_OTH, MAL\_OTH) show lower sensitivity and Dice, reflecting the difficulty of learning stable decision boundaries under data scarcity. Nevertheless, the model maintains consistent precision (mean PPV 0.596) and high specificity across almost all categories, indicating confident and reliable predictions. Overall, these results show that the weighted-fusion design generalizes well across diverse lesion types and provides a clinically meaningful foundation for multimodal skin lesion classification.

\newpage

\section{Grad-CAM Visualizations}

\begin{figure}[!h]
\centering
\includegraphics[width=0.7\linewidth]{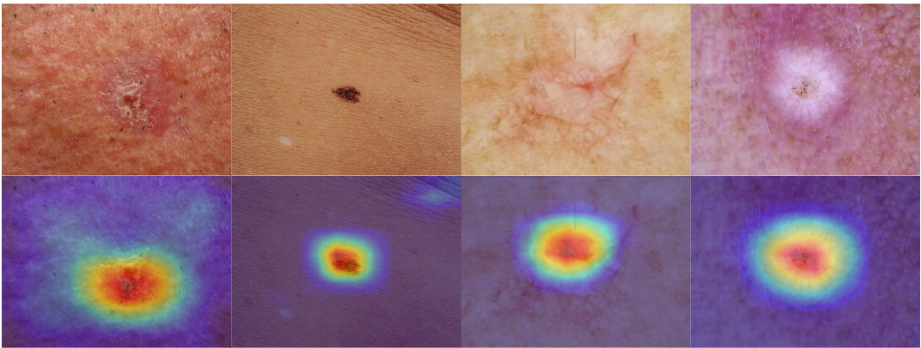}
\caption{Comparison between the original input images and their corresponding Grad-CAM visualizations. Warmer colors (yellow–red) indicate regions with higher model attention, while cooler colors (blue–green) represent lower emphasis.}
\label{fig:heat}
\end{figure}

To better understand how the proposed model interprets lesion patterns, we generate Grad-CAM visualizations for the test images. As in Figure~\ref{fig:heat}, the resulting heatmaps consistently concentrate on the main lesion regions. This indicates that the model bases its predictions on visually meaningful cues such as pigment distribution, localized texture variations, and boundary characteristics. The attention patterns are well aligned with areas that dermatologists typically examine, suggesting that the multimodal fusion framework encourages the network to focus on diagnostically relevant features. These qualitative visualizations provide additional insight into the model’s decision process and support the interpretability of the proposed approach.

\section{Calibration Curve Analysis}

\begin{figure}[!ht]
\centering
\includegraphics[width=0.4\linewidth]{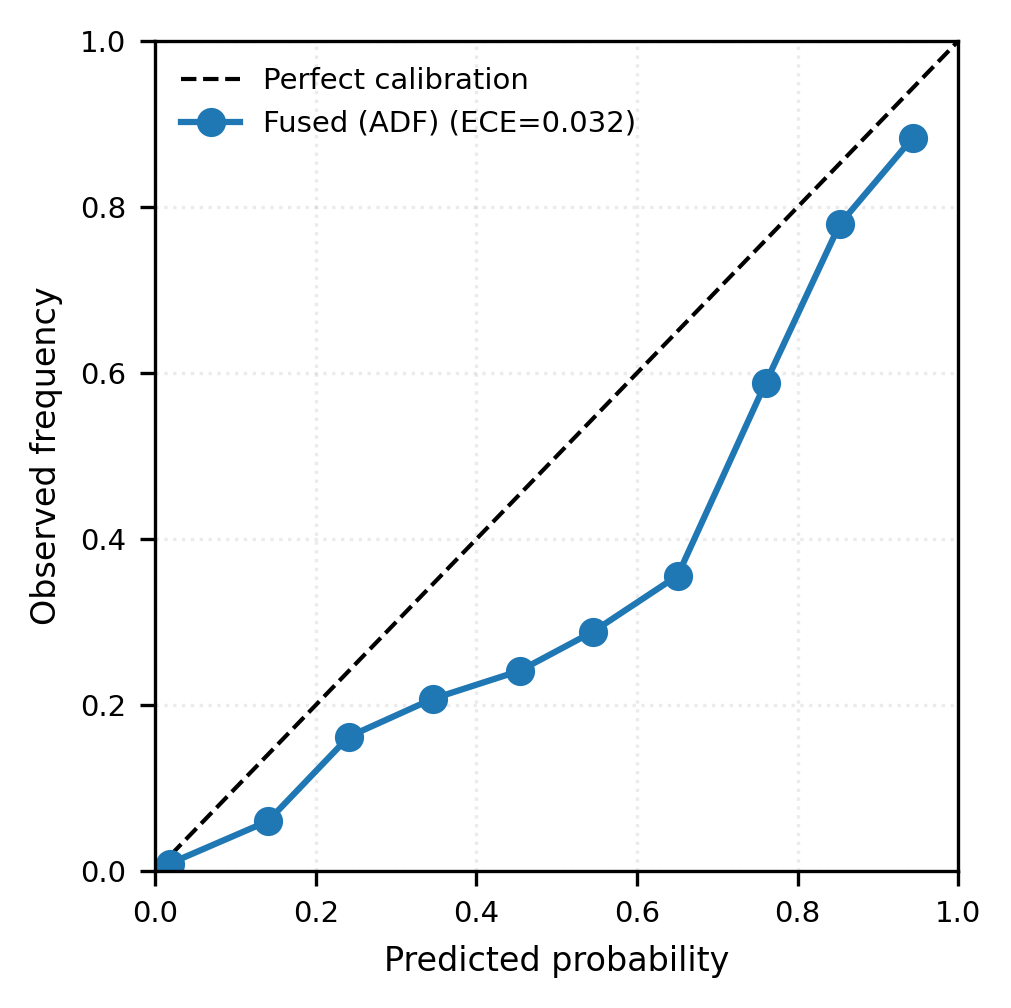}
\caption{Calibration Curve.}
\label{fig:cali}
\end{figure}

The calibration curve of the fused JI-ADF model lies close to the diagonal, indicating that predicted probabilities match observed frequencies well overall. The curve is slightly below the perfect-calibration line for mid-range probabilities, suggesting mild over-confidence in this region, but it aligns closely with the diagonal for high-confidence predictions $(\geq 0.7)$, where clinical decisions are most critical. The low expected calibration error (ECE = 0.032) confirms that the model is well calibrated globally.
%%%%%%%%%%%%%%%%%%%%%%%%%%%%%%%%%%%%%%%%%%%%%%%%%%%%%%%%%%%%%%%%%%%%%%%%%%%%%%%
%%%%%%%%%%%%%%%%%%%%%%%%%%%%%%%%%%%%%%%%%%%%%%%%%%%%%%%%%%%%%%%%%%%%%%%%%%%%%%%

\end{document}